\documentclass[runningheads]{llncs}
\usepackage{graphicx}
\usepackage{amsmath}
\usepackage{amsfonts}
\usepackage{multirow}
\usepackage{color}
\usepackage{hyperref}
\begin{document}
\title{Miss the Point: Targeted Adversarial Attack on Multiple Landmark Detection}
\titlerunning{Targeted Adversarial Attack on Multiple Landmark Detection}
%
\author{Qingsong Yao\inst{1, 3} \and
Zecheng He\inst{2} \and
Hu Han\inst{1, 3} \and
S. Kevin Zhou\inst{1, 3}
\thanks{This work is supported in part by the Youth Innovation Promotion Association CAS (grant 2018135) and Alibaba Group through Alibaba Innovative Research Program.}}

\authorrunning{Q. Yao, et al.}

\institute{\
Medical Imaging, Robotics, Analytic Computing Laboratory/Engineering (MIRACLE), Key Lab of Intelligent Information Processing of Chinese Academy of Sciences (CAS), Institute of Computing Technology, CAS, Beijing 100190, China\\
 \email{\{yaoqingsong19\}@mails.ucas.edu.cn} \email{\{hanhu,zhoushaohua\}@ict.ac.cn}  \and
Princeton University \\ \email{zechengh@princeton.edu} \and
Peng Cheng Laboratory, Shenzhen, China}
\maketitle              

\begin{abstract}
Recent methods in multiple landmark detection based on deep convolutional neural networks (CNNs) reach high accuracy and improve traditional clinical workflow. However, the vulnerability of CNNs to adversarial-example attacks can be easily exploited to break classification and segmentation tasks. This paper is the first to study how fragile a CNN-based model on multiple landmark detection to adversarial perturbations. Specifically, we propose a novel Adaptive Targeted Iterative FGSM (ATI-FGSM) attack against the state-of-the-art models in multiple landmark detection. The attacker can use ATI-FGSM to precisely control the model predictions of arbitrarily selected landmarks, while keeping other stationary landmarks still, by adding imperceptible perturbations to the original image. A comprehensive evaluation on a public dataset for cephalometric landmark detection demonstrates that the adversarial examples generated by ATI-FGSM break the CNN-based network more effectively and efficiently, compared with the original Iterative FGSM attack. Our work reveals serious threats to patients' health. Furthermore, we discuss the limitations of our method and provide potential defense directions, by investigating the coupling effect of nearby landmarks, i.e., a major source of divergence in our experiments. Our source code is available at \href{https://github.com/qsyao/attack\_landmark\_detection}{https://github.com/qsyao/attack\_landmark\_detection}.

\end{abstract}
\section{Introduction}

Multiple landmark detection is an important pre-processing step in therapy planning and intervention, thus it has attracted great interest from academia and industry \cite{zhou2015medical,zhou2017deep,5540016,yang2017automatic}. It has been successfully applied to many practical medical clinical scenarios such as knee joint surgery \cite{yang2015automated}, orthognathic and maxillofacial surgeries \cite{chen2019cephalometric}, carotid artery bifurcation \cite{zheng20153d}, pelvic trauma surgery \cite{bier2018x}, bone age estimation \cite{gertych2007bone}. Also, it is an important step in medical imaging analysis \cite{payer2016regressing,litjens2017survey,9058664}, e.g.,  registration or initialization of segmentation algorithms.

Recently, CNN-based methods has rapidly become a methodology of choice for analyzing medical images. Compared with expert manual annotation, CNN achieves high accuracy and efficiency at a low-cost \cite{chen2019cephalometric}, showing great potential in multiple landmark detection. 
Chen et al. \cite{chen2019cephalometric} use cascade U-Net to launch a two-stage heatmap regression \cite{payer2016regressing}, which is widely used in medical landmark detection. Zhong et al. \cite{zhong2019attention} accomplish the task by regressing the heatmap and coordinate offset maps at the same time.

However, the vulnerability of CNNs to adversarial attacks can not be overlooked \cite{szegedy2013intriguing}. The attacks are legitimate examples with human-imperceptible perturbations, which attempt to fool a trained model to make incorrect predictions \cite{he2019non}. Goodfellow et al. \cite{goodfellow2014explaining} develop a fast gradient sign method (FGSM) to generate perturbations by back-propagating the adversarial gradient induced by an intended incorrect prediction. Kurakin et al. \cite{kurakin2016adversarial} extend it to Targeted Iterative FGSM by generating the perturbations iteratively to hack the network to predict the attacker desired target class. Adversarial attacks against CNN models become a real threat not only in classification tasks but also in segmentation and localization \cite{xie2017adversarial}. The dense adversary generation (DAG) algorithm proposed in \cite{xie2017adversarial} by Xie et al. aims to force the CNN based network to predict all pixels to target classes without $L_\infty$ norm limitation. Other works that apply the adversarial attack to classification and segmentation \cite{ozbulak2019impact,he2019non,Generalizability2018Paschali} hack the network in both targeted and non-targeted manners with a high success rate.

A targeted attack on landmark detection is stealthy and disastrous as the detection precision is tightly related to a patient's health during surgical intervention, clinical diagnosis or measurement, etc. To study the vulnerability of landmark detection systems, we propose an approach for targeted attack against CNN-based models in this paper. Our main contributions are:

\begin{figure}[t]
\centering
\includegraphics[width=1\textwidth]{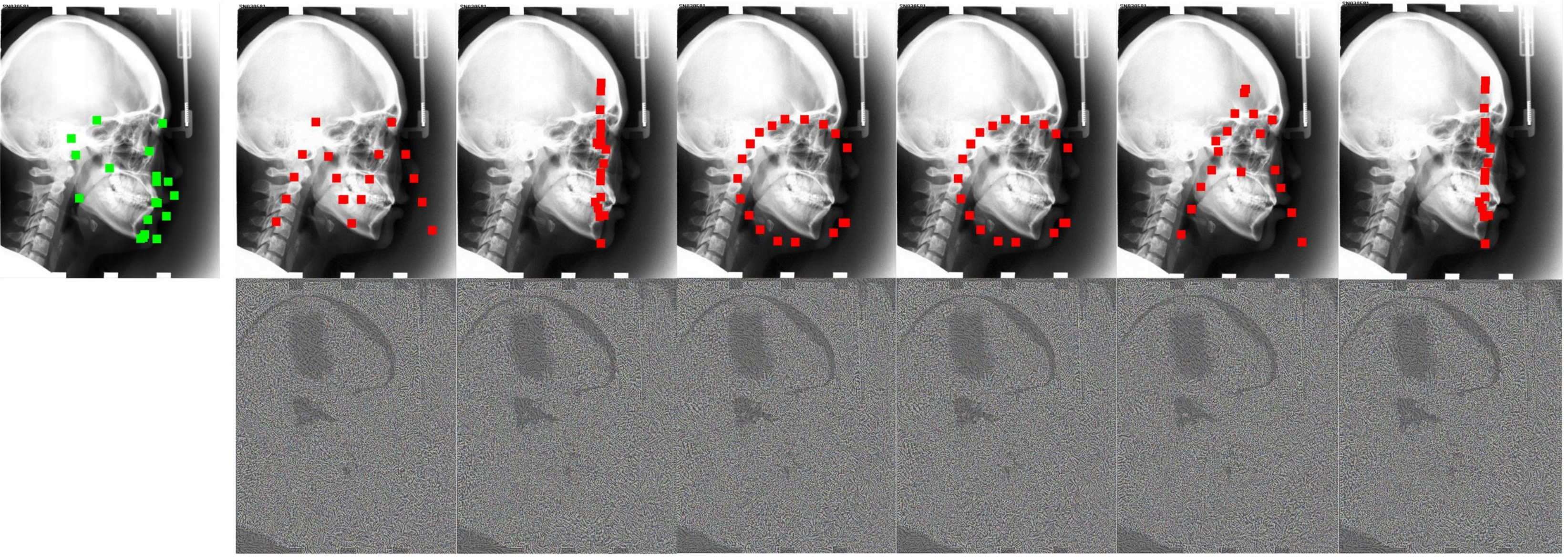}
\caption{An example of targeted adversarial attack against multiple landmark detection in a cephalometric radiograph. By adding imperceptible perturbations to the original image (left most), we arbitrarily position 19 landmarks to form the letters `MICCAI'. The perturbation is magnified by a factor of 8 for visualization.} 
\label{miccai}
\end{figure}

\begin{enumerate}
    \item A simple yet representative multi-task U-Net to detect multiple landmarks with high precision and high speed.
    \item The first targeted adversarial attack against multiple landmark detection, to the best of our knowledge, which exposes the great vulnerability of medical images against adversarial attack.
    \item An  Adaptive Targeted Iterative FGSM (ATI-FGSM) algorithm that makes the attack more effective and efficient than the standard I-FGSM.
    \item A comprehensive evaluation of the proposed algorithm to attack the landmark detection and understanding its limitations.
\end{enumerate}

\section{Multi-task U-Net for multiple landmark detection}


Existing 
approaches for multiple landmark detection use a heatmap \cite{payer2016regressing,zhong2019attention} and/or  coordinate offset maps \cite{chen2019cephalometric} to represent a landmark and then a U-Net-like network \cite{ronneberger2015u} is learned to predict the above map(s), which are post-processed to derive the final landmark location. Here we implement a multi-task U-Net to predict both heatmap and offset maps simultaneously and treat this network as our target model to attack.


For the $i^{th}$ landmark located at $(x_i,y_i)$ in an image $X$, its heatmap  $Y_{i}^h$ is computed as a Gaussian function $Y_{i}^h(x,y)=exp[-\frac{1}{2\sigma^2}((x-x_i)^2+(y-y_i)^2)]$ and its $x$-offset map $Y_{i}^{o_x}$ predicts the relative offset vector $Y_{i}^{o_x} = (x-x_i)/\sigma$ from $x$ to the corresponding landmark $x_i$. Similarly, its $y$-offset map $Y_{i}^{o_y}$ is defined. Different from \cite{zhong2019attention}, we truncate the map functions to zero for the pixels whose $Y_{i}^h(x,y) \ge 0.6$.
We use a binary cross-entropy loss $L^h$ to punish the divergence of predicted and ground-truth heatmaps, and an $L_1$ loss $L^o$ to punish the difference in coordinate offset maps. Here is the loss function $L_{i}$ for the $i^{th}$ landmark:
\begin{equation}
L_{i}(Y_i, g_i(X, \theta)) = \alpha L_{i}^h(Y_{i}^h, g_{i}^h(X, \theta)) + sign(Y_{i}^h)\sum_{o \in \{o_x,o_y\}}L_{i}^{o}(Y_{i}^{o}, g_{i}^{o}(X, \theta))
\label{eq:loss}
\end{equation}
where $g_{i}^h(X, \theta)$ and $g_{i}^o(X, \theta)$ are the networks that predict heatmaps and coordinate offset maps, respectively; $\theta$ is the network parameters; $\alpha$ is a balancing coefficient, and $sign(\cdot)$ is a sign function which is used to ensure that only the area highlighted by heatmap is included for calculation.

\begin{figure}[t]
\centering
\includegraphics[width=0.95\textwidth]{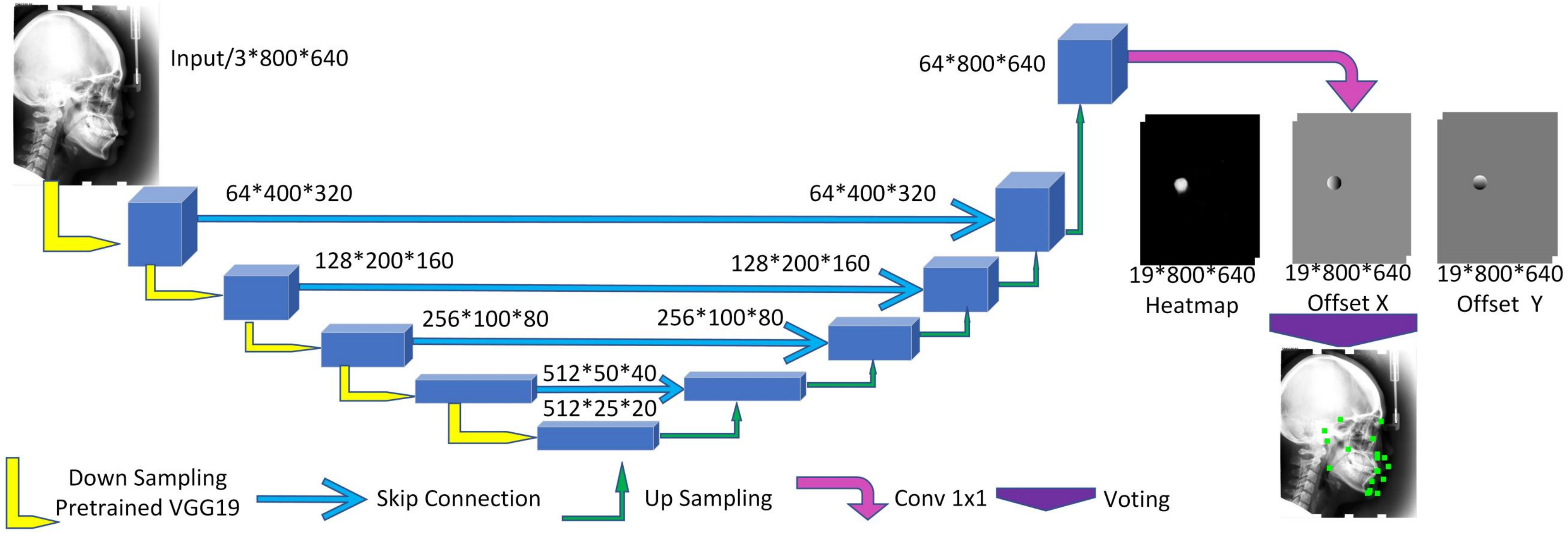}
\caption{Overview of our Multi-Task U-Net. For coordinate offset maps, we only focus on the areas with heatmap value $\ge$ 0.6. We set the value of the other areas to 0. } \label{architecture}
\end{figure}

To deal with the limited data problem, we fine-tune the encoder of U-Net initialized by the VGG19 network \cite{simonyan2014very} pretrained on ImageNet \cite{deng2009imagenet}. In the test phase, a majority-vote for candidate landmarks is conducted among all pixels with heatmap value $g_{i}^h(X, \theta)$ $\ge$ 0.6, according to their coordinate offset maps in $g_{i}^o(X, \theta)$. The winning position in the $i^{th}$ channel is the final predicted $i^{th}$ landmark \cite{chen2019cephalometric}. The whole framework is illustrated in Fig. \ref{architecture}. 

\section{Adversarial attack on multiple landmark detection}
\subsubsection{A general formulation.}
Given an original image $X_0$, the attacker attempts to generate a small perturbation $P$, such that (i) $P$ is not perceptible to human beings and (ii) for the perturbed image, i.e., $X=X_0+P$, its model prediction $g(X)$ is entirely controlled by the adversary. Follow the convention, we model the non-perceptive property of the perturbation as a constraint that the $L_\infty$ norm of $P=X-X_0$ is less than $\epsilon$. In the context of targeted adversarial attack on multiple landmark detection, taking full control of the model prediction means that, for an image with $K$ landmarks, the attacker is able to move $N$ arbitrary target landmarks to desired locations while leaving the remaining $(K-N)$ landmarks stationary. 

We denote that the index set of all landmarks is given by $\Omega=\{1,2,\ldots,K\}$ and is split into two complementary subsets: ${\cal T}=\{t_1,t_2,\ldots,t_N\}$ for the indices of target landmarks and ${\cal S}=\{s_1,s_2,\ldots,s_{K-N}\}$ for the indices of stationary landmarks. We model the adversarial attack against the model prediction as minimizing the Euclidean distance between any target landmark $g_t(X,\theta)$ w.r.t. its corresponding adversarial target location $(x_{t}, y_{t})$, while keeping any stationary landmark $g_s(X,\theta)$ close to its original location $(x_{s}, y_{s})$:
\begin{equation}
\begin{aligned}
\min_{X} &\sum_{t \in {\cal T}}||g_{t}(X,\theta) - (x_{t}, y_{t})||_2 + \sum_{s \in {\cal S}}||g_s(X,\theta) - (x_s, y_s)||_2\\
& s.t. \quad ||P||_\infty= ||X-X_0||_\infty \le \epsilon, X \in [0, 256]^{C\times H\times M}
\end{aligned}
\label{defination}
\end{equation}
To accommodate the map-based landmark representation, 
the attacker sets an adversarial heatmap $Y_t^h$ and coordinate offset maps $Y_t^o$ 
based on the desired position. We then replace the corresponding $Y_i^h$ and $Y_i^o$ with $Y_t^h$ and $Y_t^o$ in Eq. (\ref{eq:loss}). For a stationary landmark, we use heatmap $g^{h}(X, \theta)$ and coordinate offset maps $g^{o}(X, \theta)$ predicted by the original network as ground truth $Y_s$.
\begin{equation}
\min_{X} L(Y, g(X, \theta)) = \sum_{t \in {\cal T}}L_{t}(Y_{t}, g_{t}(X, \theta))+ \sum_{s \in {\cal S}}L_{s}(Y_{s}, g_{s}(X, \theta)).
\label{E1}
\end{equation}

\subsubsection{Targeted iterative FGSM \cite{kurakin2016adversarial}.} 
Targeted iterative FGSM is an enhanced version of Targeted FGSM \cite{kurakin2016adversarial}, increasing the attack effectiveness by iteratively tuning an adversarial example. We adapt Targeted iterative FGSM from the classification task to multiple landmark detection task by revising its loss function $L$ in Eq. (\ref{E1}). As $L$ decreases, the predicted heatmaps and coordinate offset maps converge to the adversarial ones. This moves the targeted landmark to the desired position while leaving the stationary landmarks at their original positions. The process of an adversarial example generation, i.e., decreasing $L$, is given by:
\begin{equation}
X_0^{adv} = X_0,~~
X^{adv}_{i+1}  = clip[X^{adv}_{i} - \eta \cdot sign(\bigtriangledown_{X^{adv}_{i}}L(Y, g(X^{adv}_{i}, \theta))), \epsilon]
\label{E2}
\end{equation}

\subsubsection{Adaptive targeted iterative FGSM (ATI-FGSM).}
There is a defect when directly adapting iterative FGSM from classification to landmark detection. In classification, each image is assigned a single label. However, for landmark detection, an input image contains multiple landmarks at various locations. The difficulty of moving each landmark varies significantly. Furthermore, the landmarks are not independent, thus moving one landmark may affect another. For example, moving a cohort of close landmarks (say around the jaw) to different locations at the same time is hard. To deal with this problem, we follow our intuition, that is, the relative vulnerability of each landmark to adversarial attack can be dynamically estimated based on the corresponding loss. A large loss term $L_j$ at iteration $i$ indicates that the landmark $j$ is hard to converge to the desired position at round $i$ and vice versa. Thus, we adaptively assign a weight for each landmark's loss term \textit{in each iteration}, e.g., a hard-to-converge landmark is associated with a large loss, resulting in faster and better convergence during network back-propagation. 
Formally, in each iteration, we have:
\begin{equation}
\begin{aligned}
L^{ada}(Y, g(X, \theta)) &= \sum_{t \in {\cal T}}\alpha_t\cdot L_{t}(Y_{t}, g_{t}(X, \theta))+ \sum_{s \in {\cal S}}\alpha_s\cdot L_{s}(Y_{s}, g_{s}(X, \theta))\\
\alpha_j &= L_{j} / mean(L(Y, g(X, \theta))) \quad j \in [1, K]\\
\end{aligned}
\label{adaptive loss}
\end{equation}
where $L(Y, g(X, \theta))$ is calculated by Eq. (\ref{E1}). The new $L^{ada}(Y, g(X, \theta))$, rather than the original $L(Y, g(X, \theta))$, is differentiated to generate gradient map in each iteration of our proposed ATI-FGSM attack.

\section{Experiments}
\subsubsection{Dataset and implementation details.} We use a public dataset for cephalometric landmark detection, provided in IEEE ISBI 2015 Challenge \cite{wang2016benchmark}, which contains 400 cephalometric radiographs. Each radiograph has 19 manually labeled landmarks of clinical anatomical significance by the two expert doctors. We take the average annotations by two doctors as the ground truth landmarks. The image size is $1935 \times 2400$, while the pixel spacing is 0.1mm. The radiographs are split to 3 sets (Train, Test1, Test2) according to the official website, whose numbers of images are 150, 150, 100 respectively. We use mean radial error (MRE) to measure the Euclidean distance between two landmarks and successful detection rate (SDR) in four radii (2mm, 2.5mm, 3mm, 4mm), which are designated by the official challenge, to measure the performance for both adversarial attack and multi-task U-Net. As MRE can be affected by extreme values, we report median radial error (MedRE) for adversarial attacks additionally.
Our multi-task U-Net is trained on a Quadro RTX 8000 GPU and optimized by the Adam optimizer with default settings. We set $\sigma=40$. The learning rate is set to 1e-3 and decayed by 0.1 every 100 epochs. After multiple trials, we select $\alpha=1.0$ for heatmaps in Eq. (\ref{eq:loss}). We resize the input image to $800 \times 640$ and normalize the values to [-1, 1]. Finally, we train our multi-task U-Net for 230 epochs with a  batch size of 8. In the adversarial attack phase, we set $\eta$=0.05 in Eq. (\ref{E2}) for the iterative increment of perturbations in our experiments. 


\begin{table}[t]
\centering
\caption{Comparison of five state-of-the-art methods and our proposed multi-task U-Net on the IEEE ISBI 2015 Challenge \cite{wang2016benchmark} datasets. We use the proposed multi-task U-Net as the target model to hack.}
\begin{tabular}{|l|ccccc|ccccc}
\hline
\multirow{2}{*}{Model} & \multicolumn{5}{c|}{Test Dataset 1} & \multicolumn{5}{c|}{Test Dataset 2} \\ \cline{2-11} 
 & MRE & 2mm & 2.5mm & 3mm & 4mm & MRE & 2mm & 2.5mm & 3mm & \multicolumn{1}{c|}{4mm} \\ \hline
Ibragimov et al. \cite{ibragimov2015computerized} & 1.87 & 71.70 & 77.40 & 81.90 & 88.00 & - & 62.74 & 70.47 & 76.53 & \multicolumn{1}{c|}{85.11} \\
Lindner et al. \cite{lindner2015fully} & 1.67 & 74.95 & 80.28 & 84.56 & 89.68 & - & 66.11 & 72.00 & 77.63 & \multicolumn{1}{c|}{87.42} \\
Arik et al. \cite{arik2017fully} & - & 75.37 & 80.91 & 84.32 & 88.25 & - & 67.68 & 74.16 & 79.11 & \multicolumn{1}{c|}{84.63} \\
Zhong et al. \cite{zhong2019attention} & 1.14 & 86.74 & 92.00 & 94.71 & 97.82 & - & - & - & - & \multicolumn{1}{c|}{-} \\
Chen et al. \cite{chen2019cephalometric} & 1.17 & 86.67 & 92.67 & 95.54 & 98.53 & 1.48 & 75.05 & 82.84 & 88.53 & \multicolumn{1}{c|}{95.05} \\
Proposed & 1.24 & 84.84 & 90.52 & 93.75 & 97.40 & 1.61 & 71.89 & 80.63 & 86.36 & \multicolumn{1}{c|}{93.68} \\ \hline
\end{tabular}
\label{Unet}
\end{table}

\textbf{Detection performance of multi-task U-Net.} 
We report the performance of our multi-task U-Net and compare it with five state-of-the-art methods \cite{ibragimov2015computerized,lindner2015fully,arik2017fully,zhong2019attention,chen2019cephalometric} in Table \ref{Unet}. Our proposed approach predicts the positions of the landmarks only by regressing heatmaps and coordinate offset maps, which are widely used in the landmark detection task \cite{payer2016regressing,zhong2019attention,chen2019cephalometric}. In terms of performance, our approach is close to the state-of-the-art methods \cite{chen2019cephalometric,zhong2019attention} and significantly ahead of the IEEE ISBI 2015 Challenge championship \cite{lindner2015fully}.

\begin{figure}[t]
\centering
\includegraphics[width=0.8\textwidth]{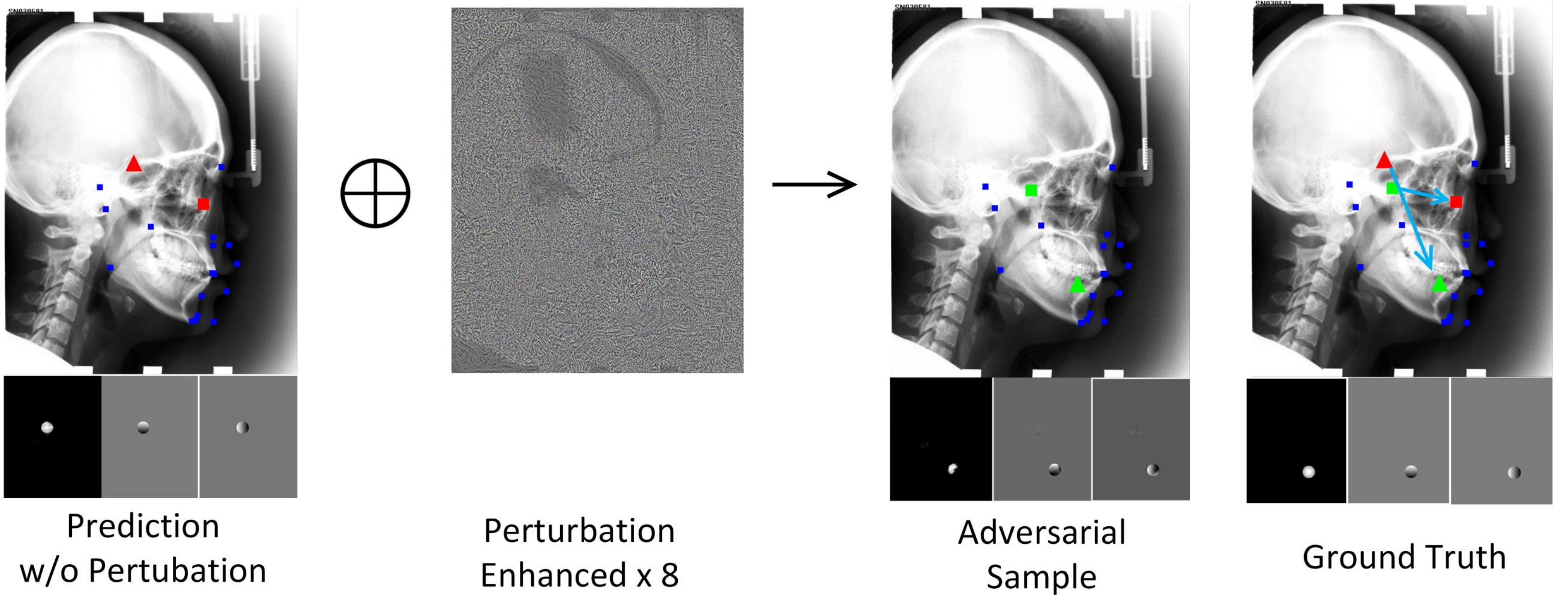}
\caption{An example of a targeted two-landmarks adversarial attack using our proposed Adaptive Targeted Iterative FGSM (ATI-FGSM). Red points highlight the predicted landmarks by the model on the original image, which are very close to the ground truth landmarks. After adding imperceptible perturbation to the input image, the model predicts the green points as the corresponding landmark positions. The green points are far away from their original positions (red), and can be controlled by the adversary. On the other hand, all other stationary points (blue) remain close to their original positions.}
\label{attack}
\end{figure}

\subsection{Performance of ATI-FSGM}

We evaluate the performance of the ATI-FGSM attack against the multi-task U-Net. To simulate the hardest scenario, our evaluation is established in a completely random setting. For each raw image in the two test datasets (250 images in total), we repeat twice the following: First randomly select a number of landmarks as targeted landmarks, leaving the rest as stationary landmarks. Then the target coordinates are randomly generated for the selected landmarks, from a huge rectangle ($x\in[100, 600], y\in[250, 750]$). So we have 500 attack attempts in total. This high level of randomness introduces significantly difficult cases for the adversarial attack. We generate adversarial examples by iterating 300 times (unless otherwise specified), under the constraint that $\epsilon=8$. 
As in Fig. \ref{attack}, the adversarial example moves the targeted landmarks (red) to the target positions (green) by fooling the network to generate incorrect heatmaps and coordinate offset maps. The small perturbation between the adversarial example and raw radiograph is hard to percept by humans. As in Table \ref{epsilon}, the goals of our adversarial attack are quickly achieved in 300 iterations, and continuously optimized for the remaining 700 iterations. Evidently, the widely used method like heatmap regression \cite{payer2016regressing} for landmark detection is very vulnerable to our attack.

\begin{table}[t]
\centering
\caption{Attack performance at different iterations and $L_\infty$ constraints. }
\begin{tabular}{|l|cccccccccccc|}
\hline
\# of iterations ($\epsilon=8$) & 1 & 20 & 50 & 100 & 150 & 200 & 250 & 300 & 400 & 600 & 750 & 1000 \\ \hline
Targeted MRE (mm) & 71.6 & 51.7 & 33.0 & 22.2 & 18.0 & 15.5 & 14.1 & 12.3 & 11.5 & 10.7 & 9.9 & 9.1 \\
Stationary MRE (mm) & 1.49 & 4.43 & 6.23 & 5.21 & 5.49 & 5.07 & 5.00 & 5.31 & 5.63 & 5.43 & 5.38 & 5.06 \\ \hline
Targeted MedRE (mm) & 69.2 & 49.6 & 1.32 & 0.67 & 0.55 & 0.42 & 0.42 & 0.42 & 0.42 & 0.42 & 0.36 & 0.33 \\
Stationary MedRE (mm) & 1.08 & 1.21 & 1.17 & 1.09 & 1.09 & 1.08 & 1.08 & 1.08 & 1.08 & 1.09 & 1.09 & 1.09 \\ \hline
Targeted 4mm SDR (\%) & 0.7 & 31.1 & 55.2 & 68.2 & 73.8 & 77.5 & 79.6 & 82.2 & 83.4 & 84.9 & 86.2 & 87.3 \\
Stationary 4mm SDR (\%) & 95.9 & 94.4 & 92.9 & 93.5 & 93.3 & 93.3 & 93.9 & 92.8 & 93.2 & 93.3 & 94.1 & 93.8 \\ \hline
$\epsilon$-value (\# of iterations=300) & - & - & - & 0.5 & 1 & 2 & 4 & 8 & 16 & 32 & - & - \\ \hline
Targeted MRE (mm) & - & - & - & 71.1 & 65.0 & 43.3 & 22.7 & 12.3 & 9.0 & 7.9 & - & -  \\
Stationary MRE (mm) & - & - & - & 1.72 & 1.86 & 3.10 & 6.86 & 5.31 & 5.73 & 5.72 & - & - \\ \hline
Targeted MedRE (mm) & - & - & - & 72.2 & 65.1 & 34.1 & 0.5 & 0.42 & 0.42 & 0.42 & - & - \\
Stationary MedRE (mm) & - & - & - & 1.05 & 1.08 & 1.08 & 1.09 & 1.08 & 1.09 & 1.08 & - & - \\ \hline
Targeted 4mm SDR (\%) & - & - & - & 1.17 & 9.69 & 38.3 & 68.5 & 82.2 & 87.3 & 88.2 & - & - \\
Stationary 4mm SDR (\%) & - & - & - & 95.4 & 95.1 & 94.3 & 93.1 & 92.8 & 92.9 & 92.1 & - & - \\ \hline
\end{tabular}
\label{epsilon}
\end{table}

\textbf{Attack performance vs perturbation strength.} We evaluate our method under different constraints of perturbation intensity. Table \ref{epsilon} shows that the adversarial examples generated by our method can achieve low MedRE and high 4mm SDR by attacking randomly targeted landmarks successfully, while keeping most of the stationary landmarks at their original positions. Moreover, as the $L_\infty$ norm constraint relaxes, MRE drops rapidly with more difficult landmarks hacked, but a few stationary landmarks are moved away.

\begin{figure}[t]
\centering
\includegraphics[width=0.75\textwidth]{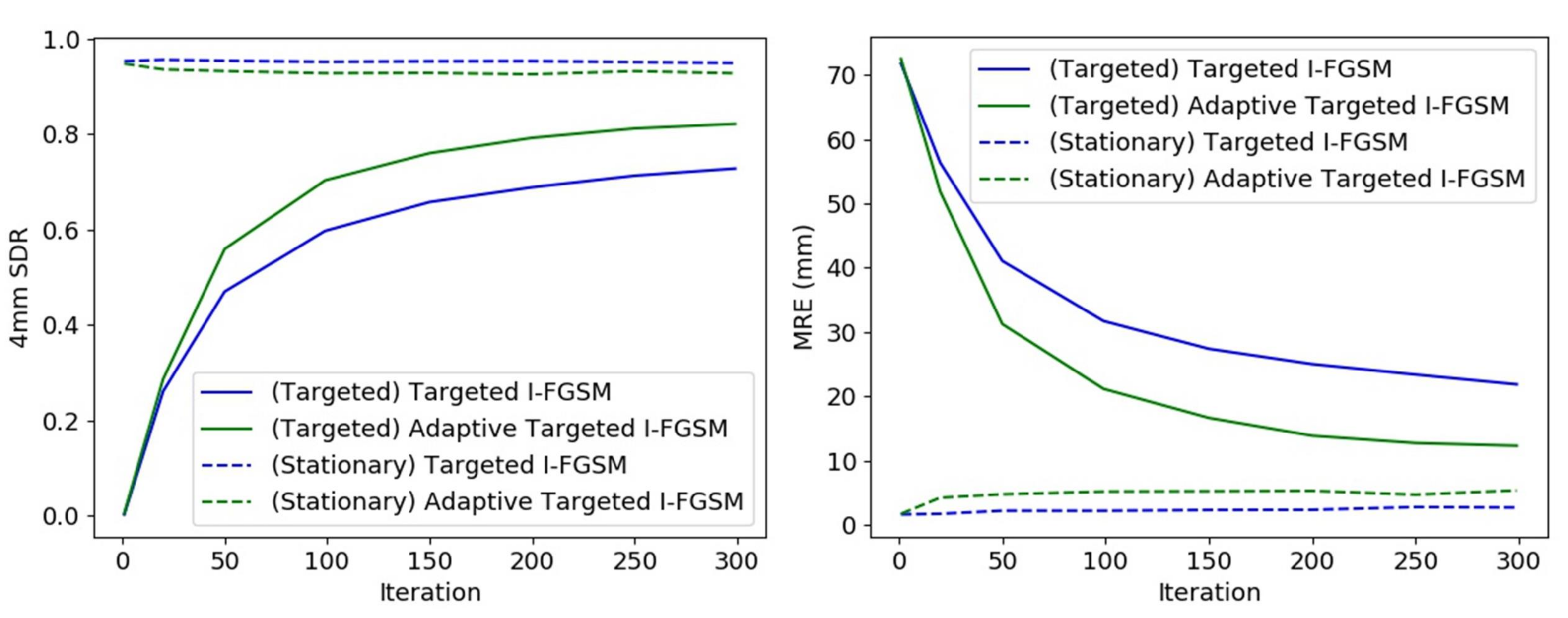}
\caption{Comparison between TI-FGSM and our ATI-FGSM. Our ATI-FGSM can arbitrarily move targeted landmarks away more efficiently and keep most of stationary landmarks in their original positions.} 
\label{mre}
\end{figure}

\textbf{The effect of adaptiveness in ATI-FGSM.} We compare the evaluation metrics and convergence speed of our method (green line) against Targeted I-FGSM (blue line) by generating 500 random adversarial examples with $\epsilon=8$. The MRE and 4mm SDR (at 300 iterations) of our method are 12.28mm and 82\% while Targeted I-FGSM convergences to 21.84mm and 72\%, respectively. Besides, our method compromises the network more quickly, results in a shorter attack time. The results in Fig. \ref{mre} show the advantages of our method lie in not only the attack effectiveness but also efficiency. Note that the attacker can not keep all of the stationary landmarks still, a few stationary landmarks are moved away by our method.


\begin{figure}[t]
\centering
\includegraphics[width=0.75\textwidth]{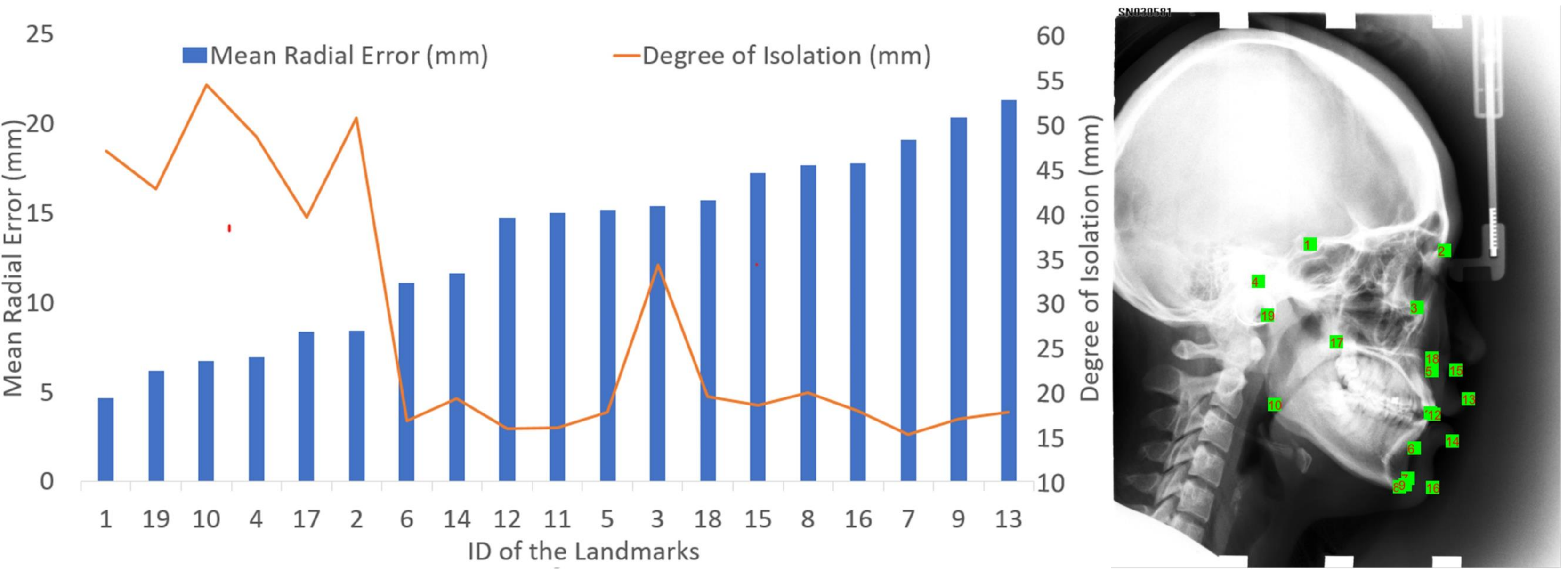}
\caption{Relationship between attack performance (measured by MRE) of each of 19 landmarks and its degree of isolation.} 
\label{failure}
\end{figure}

\textbf{Attack performance vs degree of isolation.} As some landmarks are closely related, such as landmarks on the chin or nose, which are adjacent in all images, we set up an experiment to investigate the relationship between MRE and the degree of isolation. We define the degree of isolation of a landmark as the average distance between its five nearest neighbors. As shown in Fig. \ref{failure}, we observe that MRE and degree of isolation are negatively correlated. Therefore, moving a couple of adjacent landmarks to random positions is more difficult than moving isolated ones. This is the major source of target deviation in our experiment, which may lead to potential defense. 


\textbf{A fancy attack.} We draw `MICCAI' on the same radiograph by attacking all of the 19 landmarks to the targeted position with $\epsilon=8$ and 3000 iterations, which take 600s (per image) to compute on the GPU. As in Fig. \ref{miccai}, most landmarks are hacked successfully, which justifies that the CNN-based landmark detection is vulnerable to adversarial attacks.

%

\section{Conclusion}
We demonstrate vulnerability of CNN-based multiple landmark detection models when facing the adversarial-example attack. We show that the attacker can arbitrarily manipulate landmark predictions by adding imperceptible perturbations to the original image. Furthermore, we propose the adaptive targeted iterative FGSM, a novel algorithm to launch the adversarial attack more efficiently and effectively. At last, we investigate the relationship between vulnerability and coupling of landmarks, which can be helpful in future defense.

%
%
%


\begin{thebibliography}{10}
\providecommand{\url}[1]{\texttt{#1}}
\providecommand{\urlprefix}{URL }
\providecommand{\doi}[1]{https://doi.org/#1}

\bibitem{arik2017fully}
Arik, S.{\"O}., Ibragimov, B., Xing, L.: Fully automated quantitative
  cephalometry using convolutional neural networks. Journal of Medical Imaging
  \textbf{4}(1),  014501 (2017)

\bibitem{bier2018x}
Bier, B., Unberath, M., Zaech, J.N., Fotouhi, J., Armand, M., Osgood, G.,
  Navab, N., Maier, A.: X-ray-transform invariant anatomical landmark detection
  for pelvic trauma surgery. In: MICCAI. pp. 55--63. Springer (2018)

\bibitem{chen2019cephalometric}
Chen, R., Ma, Y., Chen, N., Lee, D., Wang, W.: Cephalometric landmark detection
  by attentive feature pyramid fusion and regression-voting. In: MICCAI. pp.
  873--881. Springer (2019)

\bibitem{deng2009imagenet}
Deng, J., Dong, W., Socher, R., Li, L.J., Li, K., Fei-Fei, L.: Imagenet: A
  large-scale hierarchical image database. In: CVPR. pp. 248--255 (2009)

\bibitem{gertych2007bone}
Gertych, A., Zhang, A., Sayre, J., Pospiech-Kurkowska, S., Huang, H.: Bone age
  assessment of children using a digital hand atlas. Computerized Medical
  Imaging and Graphics  \textbf{31}(4-5),  322--331 (2007)

\bibitem{goodfellow2014explaining}
Goodfellow, I., Shlens, J., Szegedy, C.: Explaining and harnessing adversarial
  examples. In: ICLR (2015)

\bibitem{he2019non}
He, X., Yang, S., Li, G., Li, H., Chang, H., Yu, Y.: Non-local context encoder:
  Robust biomedical image segmentation against adversarial attacks. In: AAAI.
  vol.~33, pp. 8417--8424 (2019)

\bibitem{ibragimov2015computerized}
Ibragimov, B., Likar, B., Pernus, F., Vrtovec, T.: Computerized cephalometry by
  game theory with shape-and appearance-based landmark refinement (2015)

\bibitem{kurakin2016adversarial}
Kurakin, A., Goodfellow, I., Bengio, S.: Adversarial machine learning at scale.
  ICLR  (2017)

\bibitem{9058664}
{Li}, H., {Han}, H., {Li}, Z., {Wang}, L., {Wu}, Z., {Lu}, J., {Zhou}, S.K.:
  High-resolution chest x-ray bone suppression using unpaired ct structural
  priors. IEEE Transactions on Medical Imaging  (2020)

\bibitem{lindner2015fully}
Lindner, C., Cootes, T.F.: Fully automatic cephalometric evaluation using
  random forest regression-voting. Scientific Reports  \textbf{6},  33581
  (2016)

\bibitem{litjens2017survey}
Litjens, G., Kooi, T., Bejnordi, B.E., Setio, A.A.A., Ciompi, F., Ghafoorian,
  M., Van Der~Laak, J.A., Van~Ginneken, B., S{\'a}nchez, C.I.: A survey on deep
  learning in medical image analysis. Medical Image Analysis  \textbf{42},
  60--88 (2017)

\bibitem{5540016}
{Liu}, D., {Zhou}, S.K., {Bernhardt}, D., {Comaniciu}, D.: Search strategies
  for multiple landmark detection by submodular maximization. In: CVPR. pp.
  2831--2838 (2010)

\bibitem{ozbulak2019impact}
Ozbulak, U., Van~Messem, A., De~Neve, W.: Impact of adversarial examples on
  deep learning models for biomedical image segmentation. In: MICCAI. pp.
  300--308. Springer (2019)

\bibitem{Generalizability2018Paschali}
Paschali, M., Conjeti, S., Navarro, F., Navab, N.: Generalizability vs.
  robustness: Investigating medical imaging networks using adversarial
  examples. In: MICCAI. pp. 493--501. Springer (2018)

\bibitem{payer2016regressing}
Payer, C., {\v{S}}tern, D., Bischof, H., Urschler, M.: Regressing heatmaps for
  multiple landmark localization using cnns. In: MICCAI. pp. 230--238. Springer
  (2016)

\bibitem{ronneberger2015u}
Ronneberger, O., Fischer, P., Brox, T.: U-net: Convolutional networks for
  biomedical image segmentation. In: MICCAI. pp. 234--241. Springer (2015)

\bibitem{simonyan2014very}
Simonyan, K., Zisserman, A.: Very deep convolutional networks for large-scale
  image recognition. In: ICLR (2015)

\bibitem{szegedy2013intriguing}
Szegedy, C., Zaremba, W., Sutskever, I., Bruna, J., Erhan, D., Goodfellow, I.,
  Fergus, R.: Intriguing properties of neural networks. In: ICLR (2014)

\bibitem{wang2016benchmark}
Wang, C.W., Huang, C.T., Lee, J.H., Li, C.H., Chang, S.W., Siao, M.J., Lai,
  T.M., Ibragimov, B., Vrtovec, T., Ronneberger, O., et~al.: A benchmark for
  comparison of dental radiography analysis algorithms. Medical Image Analysis
  \textbf{31},  63--76 (2016)

\bibitem{xie2017adversarial}
Xie, C., Wang, J., Zhang, Z., Zhou, Y., Xie, L., Yuille, A.: Adversarial
  examples for semantic segmentation and object detection. In: CVPR. pp.
  1369--1378 (2017)

\bibitem{yang2017automatic}
Yang, D., Xiong, T., Xu, D., Huang, Q., Liu, D., Zhou, S.K., Xu, Z., Park, J.,
  Chen, M., Tran, T.D., et~al.: Automatic vertebra labeling in large-scale 3d
  ct using deep image-to-image network with message passing and sparsity
  regularization. In: IPMI. pp. 633--644 (2017)

\bibitem{yang2015automated}
Yang, D., Zhang, S., Yan, Z., Tan, C., Li, K., Metaxas, D.: Automated
  anatomical landmark detection ondistal femur surface using convolutional
  neural network. In: ISBI. pp. 17--21 (2015)

\bibitem{zheng20153d}
Zheng, Y., Liu, D., Georgescu, B., Nguyen, H., Comaniciu, D.: 3d deep learning
  for efficient and robust landmark detection in volumetric data. In: MICCAI.
  pp. 565--572. Springer (2015)

\bibitem{zhong2019attention}
Zhong, Z., Li, J., Zhang, Z., Jiao, Z., Gao, X.: An attention-guided deep
  regression model for landmark detection in cephalograms. In: MICCAI. pp.
  540--548. Springer (2019)

\bibitem{zhou2015medical}
Zhou, S.K. (ed.): Medical image recognition, segmentation and parsing: machine
  learning and multiple object approaches. Academic Press (2015)

\bibitem{zhou2017deep}
Zhou, S.K., Greenspan, H., Shen, D. (eds.): Deep learning for medical image
  analysis. Academic Press (2017)

\end{thebibliography}
\end{document}